# SQuAD: 100,000+ Questions for Machine Comprehension of Text

**Pranav Rajpurkar** and **Jian Zhang** and **Konstantin Lopyrev** and **Percy Liang**
{pranavsr,zjian,klopyrev,pliang}@cs.stanford.edu
Computer Science Department
Stanford University

## Abstract

We present the Stanford Question Answering Dataset (SQuAD), a new reading comprehension dataset consisting of 100,000+ questions posed by crowdworkers on a set of Wikipedia articles, where the answer to each question is a segment of text from the corresponding reading passage. We analyze the dataset to understand the types of reasoning required to answer the questions, leaning heavily on dependency and constituency trees. We build a strong logistic regression model, which achieves an F1 score of 51.0%, a significant improvement over a simple baseline (20%). However, human performance (86.8%) is much higher, indicating that the dataset presents a good challenge problem for future research. The dataset is freely available at https://stanford-qa.com.

In meteorology, precipitation is any product of the condensation of atmospheric water vapor that falls under **gravity**. The main forms of precipitation include drizzle, rain, sleet, snow, **graupel** and hail... Precipitation forms as smaller droplets coalesce via collision with other rain drops or ice crystals **within a cloud**. Short, intense periods of rain in scattered locations are called "showers".

What causes precipitation to fall?
**gravity**

What is another main form of precipitation besides drizzle, rain, snow, sleet and hail?
**graupel**

Where do water droplets collide with ice crystals to form precipitation?
**within a cloud**

Figure 1: Question-answer pairs for a sample passage in the SQuAD dataset. Each of the answers is a segment of text from the passage.

## 1 Introduction

Reading Comprehension (RC), or the ability to read text and then answer questions about it, is a challenging task for machines, requiring both understanding of natural language and knowledge about the world. Consider the question "*what causes precipitation to fall?*" posed on the passage in Figure 1. In order to answer the question, one might first locate the relevant part of the passage "*precipitation ... falls under gravity*", then reason that "*under*" refers to a cause (not location), and thus determine the correct answer: "*gravity*".

How can we get a machine to make progress on the challenging task of reading comprehension? Historically, large, realistic datasets have played a critical role for driving fields forward—famous examples include ImageNet for object recognition (Deng et al., 2009) and the Penn Treebank for syntactic parsing (Marcus et al., 1993). Existing datasets for RC have one of two shortcomings: (i) those that are high in quality (Richardson et al., 2013; Berant et al., 2014) are too small for training modern data-intensive models, while (ii) those that are large (Hermann et al., 2015; Hill et al., 2015) are semi-synthetic and do not share the same characteristics as explicit reading comprehension questions.

To address the need for a large and high-quality reading comprehension dataset, we present the Stan-ford Question Answering Dataset v1.0 (SQuAD), freely available at https://stanford-qa.com, consisting of questions posed by crowdworkers on a set of Wikipedia articles, where the answer to every question is a segment of text, or *span*, from the corresponding reading passage. SQuAD contains 107,785 question-answer pairs on 536 articles, and is almost two orders of magnitude larger than previous manually labeled RC datasets such as MCTest (Richardson et al., 2013).

In contrast to prior datasets, SQuAD does not provide a list of answer choices for each question. Rather, systems must select the answer from all possible spans in the passage, thus needing to cope with a fairly large number of candidates. While questions with span-based answers are more constrained than the more interpretative questions found in more advanced standardized tests, we still find a rich diversity of questions and answer types in SQuAD. We develop automatic techniques based on distances in dependency trees to quantify this diversity and stratify the questions by difficulty. The span constraint also comes with the important benefit that span-based answers are easier to evaluate than free-form answers.

To assess the difficulty of SQuAD, we implemented a logistic regression model with a range of features. We find that lexicalized and dependency tree path features are important to the performance of the model. We also find that the model performance worsens with increasing complexity of (i) answer types and (ii) syntactic divergence between the question and the sentence containing the answer; interestingly, there is no such degradation for humans. Our best model achieves an F1 score of 51.0%,[1] which is much better than the sliding window baseline (20%). Over the last four months (since June 2016), we have witnessed significant improvements from more sophisticated neural network-based models. For example, Wang and Jiang (2016) obtained 70.3% F1 on SQuAD v1.1 (results on v1.0 are similar). These results are still well behind human performance, which is 86.8% F1 based on inter-annotator agreement. This suggests that there is plenty of room for advancement in modeling and learning on the SQuAD dataset.

---

[1] All experimental results in this paper are on SQuAD v1.0.

| Dataset | Question source | Formulation | Size |
| --- | --- | --- | --- |
| **SQuAD** | **crowdsourced** | **RC, spans in passage** | **100K** |
| MCTest (Richardson et al., 2013) | crowdsourced | RC, multiple choice | 2640 |
| Algebra (Kushman et al., 2014) | standardized tests | computation | 514 |
| Science (Clark and Etzioni, 2016) | standardized tests | reasoning, multiple choice | 855 |
| WikiQA (Yang et al., 2015) | query logs | IR, sentence selection | 3047 |
| TREC-QA (Voorhees and Tice, 2000) | query logs + human editor | IR, free form | 1479 |
| CNN/Daily Mail (Hermann et al., 2015) | summary + cloze | RC, fill in single entity | 1.4M |
| CBT (Hill et al., 2015) | cloze | RC, fill in single word | 688K |

Table 1: A survey of several reading comprehension and question answering datasets. SQuAD is much larger than all datasets except the semi-synthetic cloze-style datasets, and it is similar to TREC-QA in the open-endedness of the answers.

## 2 Existing Datasets

We begin with a survey of existing reading comprehension and question answering (QA) datasets, highlighting a variety of task formulation and creation strategies (see Table 1 for an overview).

**Reading comprehension.** A data-driven approach to reading comprehension goes back to Hirschman et al. (1999), who curated a dataset of 600 real 3rd–6th grade reading comprehension questions. Their pattern matching baseline was subsequently improved by a rule-based system (Riloff and Thelen, 2000) and a logistic regression model (Ng et al., 2000). More recently, Richardson et al. (2013) curated MCTest, which contains 660 stories created by crowdworkers, with 4 questions per story and 4 answer choices per question. Because many of the questions require commonsense reasoning and reasoning across multiple sentences, the dataset remains quite challenging, though there has been noticeable progress (Narasimhan and Barzilay, 2015; Sachan et al., 2015; Wang et al., 2015). Both curated datasets, although real and difficult, are too small to support very expressive statistical models.

Some datasets focus on deeper reasoning abilities. Algebra word problems require understanding a story well enough to turn it into a system of equa-tions, which can be easily solved to produce the answer (Kushman et al., 2014; Hosseini et al., 2014). BAbI (Weston et al., 2015), a fully synthetic RC dataset, is stratified by different types of reasoning required to solve each task. Clark and Etzioni (2016) describe the task of solving 4th grade science exams, and stress the need to reason with world knowledge.

**Open-domain question answering.** The goal of open-domain QA is to answer a question from a large collection of documents. The annual evaluations at the Text REtreival Conference (TREC) (Voorhees and Tice, 2000) led to many advances in open-domain QA, many of which were used in IBM Watson for Jeopardy! (Ferrucci et al., 2013). Recently, Yang et al. (2015) created the WikiQA dataset, which, like SQuAD, use Wikipedia passages as a source of answers, but their task is sentence selection, while ours requires selecting a specific span in the sentence.

Selecting the span of text that answers a question is similar to *answer extraction*, the final step in the open-domain QA pipeline, methods for which include bootstrapping surface patterns (Ravichandran and Hovy, 2002), using dependency trees (Shen and Klakow, 2006), and using a factor graph over multiple sentences (Sun et al., 2013). One key difference between our RC setting and answer extraction is that answer extraction typically exploits the fact that the answer occurs in multiple documents (Brill et al., 2002), which is more lenient than in our setting, where a system only has access to a single reading passage.

**Cloze datasets.** Recently, researchers have constructed cloze datasets, in which the goal is to predict the missing word (often a named entity) in a passage. Since these datasets can be automatically generated from naturally occurring data, they can be extremely large. The Children's Book Test (CBT) (Hill et al., 2015), for example, involves predicting a blanked-out word of a sentence given the 20 previous sentences. Hermann et al. (2015) constructed a corpus of cloze style questions by blanking out entities in abstractive summaries of CNN / Daily News articles; the goal is to fill in the entity based on the original article. While the size of this dataset is impressive, Chen et al. (2016) showed that the dataset requires less reasoning than previously thought, and concluded that performance is almost saturated.

Figure 2: The crowd-facing web interface used to collect the dataset encourages crowdworkers to use their own words while asking questions.

One difference between SQuAD questions and cloze-style queries is that answers to cloze queries are single words or entities, while answers in SQuAD often include non-entities and can be much longer phrases. Another difference is that SQuAD focuses on questions whose answers are entailed by the passage, whereas the answers to cloze-style queries are merely suggested by the passage.

## 3 Dataset Collection

We collect our dataset in three stages: curating passages, crowdsourcing question-answers on those passages, and obtaining additional answers.

**Passage curation.** To retrieve high-quality articles, we used Project Nayuki's Wikipedia's internal PageRanks to obtain the top 10000 articles of English Wikipedia, from which we sampled 536 articles uniformly at random. From each of these articles, we extracted individual paragraphs, stripping away images, figures, tables, and discarding paragraphs shorter than 500 characters. The result was 23,215 paragraphs for the 536 articles covering a wide range of topics, from musical celebrities to abstract concepts. We partitioned the articles randomly into a training set (80%), a development set (10%), and a test set (10%).

**Question-answer collection.** Next, we employed crowdworkers to create questions. We used the Daemo platform (Gaikwad et al., 2015), with Amazon Mechanical Turk as its backend. Crowdworkers were required to have a 97% HIT acceptance rate, a minimum of 1000 HITs, and be located in the United States or Canada. Workers were asked to spend 4 minutes on every paragraph, and paid $9 per hour for the number of hours required to complete the article. The task was reviewed favorably by crowdworkers, receiving positive comments on Turkopticon.

On each paragraph, crowdworkers were tasked with asking and answering up to 5 questions on the content of that paragraph. The questions had to be entered in a text field, and the answers had to be highlighted in the paragraph. To guide the workers, tasks contained a sample paragraph, and examples of good and bad questions and answers on that paragraph along with the reasons they were categorized as such. Additionally, crowdworkers were encouraged to ask questions in their own words, without copying word phrases from the paragraph. On the interface, this was reinforced by a reminder prompt at the beginning of every paragraph, and by disabling copy-paste functionality on the paragraph text.

**Additional answers collection.** To get an indication of human performance on SQuAD and to make our evaluation more robust, we obtained at least 2 additional answers for each question in the development and test sets. In the secondary answer generation task, each crowdworker was shown only the questions along with the paragraphs of an article, and asked to select the shortest span in the paragraph that answered the question. If a question was not answerable by a span in the paragraph, workers were asked to submit the question without marking an answer. Workers were recommended a speed of 5 questions for 2 minutes, and paid at the same rate of $9 per hour for the number of hours required for the entire article. Over the development and test sets, 2.6% of questions were marked unanswerable by at least one of the additional crowdworkers.

| Answer type | Percentage | Example |
|---|---|---|
| Date | 8.9% | 19 October 1512 |
| Other Numeric | 10.9% | 12 |
| Person | 12.9% | Thomas Coke |
| Location | 4.4% | Germany |
| Other Entity | 15.3% | ABC Sports |
| Common Noun Phrase | 31.8% | property damage |
| Adjective Phrase | 3.9% | second-largest |
| Verb Phrase | 5.5% | returned to Earth |
| Clause | 3.7% | to avoid trivialization |
| Other | 2.7% | quietly |

Table 2: We automatically partition our answers into the following categories. Our dataset consists of large number of answers beyond proper noun entities.

## 4 Dataset Analysis

To understand the properties of SQuAD, we analyze the questions and answers in the development set. Specifically, we explore the (i) diversity of answer types, (ii) the difficulty of questions in terms of type of reasoning required to answer them, and (iii) the degree of syntactic divergence between the question and answer sentences.

**Diversity in answers.** We automatically categorize the answers as follows: We first separate the numerical and non-numerical answers. The non-numerical answers are categorized using constituency parses and POS tags generated by Stanford CoreNLP. The proper noun phrases are further split into person, location and other entities using NER tags. In Table 2, we can see dates and other numbers make up 19.8% of the data; 32.6% of the answers are proper nouns of three different types; 31.8% are common noun phrases answers; and the remaining 15.8% are made up of adjective phrases, verb phrases, clauses and other types.

**Reasoning required to answer questions.** To get a better understanding of the reasoning required to answer the questions, we sampled 4 questions from each of the 48 articles in the development set, and then manually labeled the examples with the categories shown in Table 3. The results show that all examples have some sort of lexical or syntactic divergence between the question and the answer in the passage. Note that some examples fall into more than one category.

| Reasoning | Description | Example | Percentage |
|---|---|---|---|
| Lexical variation (synonymy) | Major correspondences between the question and the answer sentence are synonyms. | Q: What is the Rankine cycle sometimes **called**? Sentence: The Rankine cycle is sometimes **referred** to as a practical Carnot cycle. | 33.3% |
| Lexical variation (world knowledge) | Major correspondences between the question and the answer sentence require world knowledge to resolve. | Q: Which **governing bodies** have veto power? Sen.: **The European Parliament and the Council of the European Union** have powers of amendment and veto during the legislative process. | 9.1% |
| Syntactic variation | After the question is paraphrased into declarative form, its syntactic dependency structure does not match that of the answer sentence even after local modifications. | Q: What Shakespeare scholar **is currently on the faculty**? Sen.: **Current faculty include** the anthropologist Marshall Sahlins, ..., Shakespeare scholar David Bevington. | 64.1% |
| Multiple sentence reasoning | There is anaphora, or higher-level fusion of multiple sentences is required. | Q: What collection does **the V&A Theatre & Performance galleries** hold? Sen.: **The V&A Theatre & Performance galleries** opened in March 2009. ... **They** hold the UK's biggest national collection of material about live performance. | 13.6% |
| Ambiguous | We don't agree with the crowdworkers' answer, or the question does not have a unique answer. | Q: What is the main goal of criminal punishment? Sen.: **Achieving crime control via incapacitation and deterrence** is a major goal of criminal punishment. | 6.1% |

Table 3: We manually labeled 192 examples into one or more of the above categories. Words relevant to the corresponding reasoning type are bolded, and the crowdsourced answer is underlined.

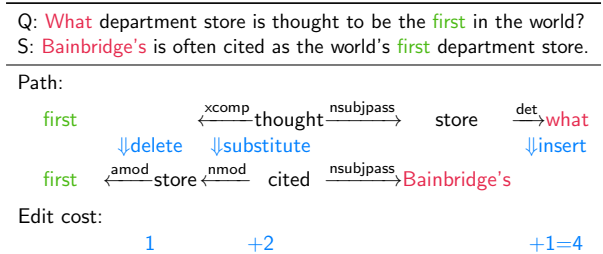

Figure 3: An example walking through the computation of the syntactic divergence between the question Q and answer sentence S.

**Stratification by syntactic divergence.** We also develop an automatic method to quantify the syntactic divergence between a question and the sentence containing the answer. This provides another way to measure the difficulty of a question and to stratify the dataset, which we return to in Section 6.3.

We illustrate how we measure the divergence with the example in Figure 3. We first detect anchors (word-lemma pairs common to both the question and answer sentences); in the example, the anchor is "*first*". The two unlexicalized paths, one from the anchor "*first*" in the question to the wh-word "*what*", and the other from the anchor in the answer sentence and to the answer span "*Bainbridge's*", are then extracted from the dependency parse trees. We measure the edit distance between these two paths, which we define as the minimum number of deletions or insertions to transform one path into the other. The syntactic divergence is then defined as the minimum edit distance over all possible anchors. The histogram in Figure 4a shows that there is a wide range of syntactic divergence in our dataset. We also show a concrete example where the edit distance is 0 and another where it is 6. Note that our syntactic divergence ignores lexical variation. Also, small divergence does not mean that a question is easy since there could be other candidates with similarly small divergence.

## 5 Methods

We developed a logistic regression model and compare its accuracy with that of three baseline methods.

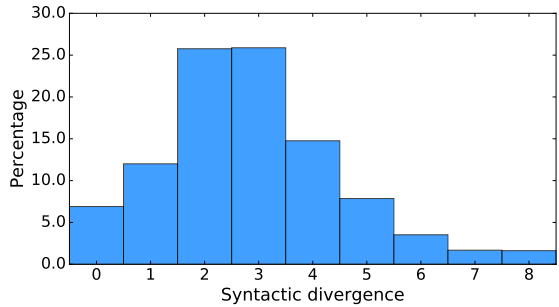
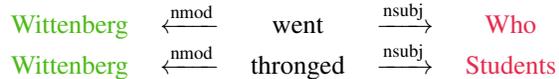

(a) Histogram of syntactic divergence.

(b) An example of a question-answer pair with edit distance 0 between the dependency paths (note that lexical variation is ignored in the computation of edit distance).

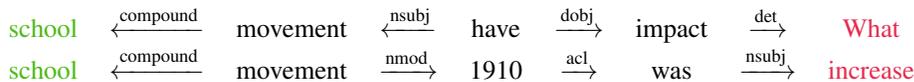

(c) An example of a question-answer pair with edit distance 6.

**Figure 4:** We use the edit distance between the unlexicalized dependency paths in the question and the sentence containing the answer to measure *syntactic divergence*.

**Candidate answer generation.** For all four methods, rather than considering all $O(L^2)$ spans as candidate answers, where $L$ is the number of words in the sentence, we only use spans which are constituents in the constituency parse generated by Stanford CoreNLP. Ignoring punctuation and articles, we find that 77.3% of the correct answers in the development set are constituents. This places an effective ceiling on the accuracy of our methods. During training, when the correct answer of an example is not a constituent, we use the shortest constituent containing the correct answer as the target.

### 5.1 Sliding Window Baseline

For each candidate answer, we compute the unigram/bigram overlap between the sentence containing it (excluding the candidate itself) and the question. We keep all the candidates that have the maximal overlap. Among these, we select the best one using the sliding-window approach proposed in Richardson et al. (2013).

In addition to the basic sliding window approach, we also implemented the distance-based extension (Richardson et al., 2013). Whereas Richardson et al. (2013) used the entire passage as the context of an answer, we used only the sentence containing the candidate answer for efficiency.

### 5.2 Logistic Regression

In our logistic regression model, we extract several types of features for each candidate answer. We discretize each continuous feature into 10 equally-sized buckets, building a total of 180 million features, most of which are lexicalized features or dependency tree path features. The descriptions and examples of the features are summarized in Table 4.

The matching word and bigram frequencies as well as the root match features help the model pick the correct sentences. Length features bias the model towards picking common lengths and positions for answer spans, while span word frequencies bias the model against uninformative words. Constituent label and span POS tag features guide the model towards the correct answer types. In addition to these basic features, we resolve lexical variation using lexicalized features, and syntactic variation using dependency tree path features.

The multiclass log-likelihood loss is optimized using AdaGrad with an initial learning rate of 0.1. Each update is performed on the batch of all questions in a paragraph for efficiency, since they share the same candidates. $L_2$ regularization is used, with a coefficient of 0.1 divided by the number of batches. The model is trained with three passes over the train-ing data.

| Feature Groups | Description | Examples |
| --- | --- | --- |
| Matching Word Frequencies | Sum of the TF-IDF of the words that occur in both the question and the sentence containing the candidate answer. Separate features are used for the words to the left, to the right, inside the span, and in the whole sentence. | Span: $[0 \leq \text{sum} < 0.01]$ <br> Left: $[7.9 \leq \text{sum} < 10.7]$ |
| Matching Bigram Frequencies | Same as above, but using bigrams. We use the generalization of the TF-IDF described in Shirakawa et al. (2015). | Span: $[0 \leq \text{sum} < 2.4]$ <br> Left: $[0 \leq \text{sum} < 2.7]$ |
| Root Match | Whether the dependency parse tree roots of the question and sentence match, whether the sentence contains the root of the dependency parse tree of the question, and whether the question contains the root of the dependency parse tree of the sentence. | Root Match = False |
| Lengths | Number of words to the left, to the right, inside the span, and in the whole sentence. | Span: $[1 <= \text{num} < 2]$ <br> Left: $[15 \leq \text{num} < 19]$ |
| Span Word Frequencies | Sum of the TF-IDF of the words in the span, regardless of whether they appear in the question. | Span: $[5.2 \leq \text{sum} < 6.9]$ |
| Constituent Label | Constituency parse tree label of the span, optionally combined with the wh-word in the question. | Span: NP <br> Span: NP, wh-word: "*what*" |
| Span POS Tags | Sequence of the part-of-speech tags in the span, optionally combined with the wh-word in the question. | Span: [NN] <br> Span: [NN], wh-word: "*what*" |
| Lexicalized | Lemmas of question words combined with the lemmas of words within distance 2 to the span in the sentence based on the dependency parse trees. Separately, question word lemmas combined with answer word lemmas. | Q: "*cause*", S: "*under*" $\xleftarrow{\text{case}}$ <br> Q: "*fall*", A: "*gravity*" |
| Dependency Tree Paths | For each word that occurs in both the question and sentence, the path in the dependency parse tree from that word in the sentence to the span, optionally combined with the path from the wh-word to the word in the question. POS tags are included in the paths. | VBZ $\xrightarrow{\text{nmod}}$ NN <br> what $\xleftarrow{\text{nsubj}}$ VBZ $\xrightarrow{\text{advcl}}$ <br> + VBZ $\xrightarrow{\text{nmod}}$ NN |

Table 4: Features used in the logistic regression model with examples for the question "*What causes precipitation to fall?*", sentence "*In meteorology, precipitation is any product of the condensation of atmospheric water vapor that falls under gravity.*" and answer "*gravity*". Q denotes question, A denotes candidate answer, and S denotes sentence containing the candidate answer.

## 6 Experiments

### 6.1 Model Evaluation

We use two different metrics to evaluate model accuracy. Both metrics ignore punctuations and articles (a, an, the).

**Exact match.** This metric measures the percentage of predictions that match any one of the ground truth answers exactly.

**(Macro-averaged) F1 score.** This metric measures the average overlap between the prediction and ground truth answer. We treat the prediction and ground truth as bags of tokens, and compute their F1. We take the maximum F1 over all of the ground truth answers for a given question, and then average over all of the questions.

### 6.2 Human Performance

We assess human performance on SQuAD's development and test sets. Recall that each of the questions in these sets has at least three answers. To evaluate human performance, we treat the second answer to each question as the human prediction, and keep the other answers as ground truth answers. The resulting human performance score on the test set is 77.0% for the exact match metric, and 86.8% for F1. Mismatch occurs mostly due to inclusion/exclusion of non-essential phrases (e.g., *monsoon trough* versus *movement of the monsoon trough*) rather than fundamental disagreements about the answer.

### 6.3 Model Performance

Table 5 shows the performance of our models alongside human performance on the v1.0 of development and test sets. The logistic regression model significantly outperforms the baselines, but underperforms humans. We note that the model is able to select the sentence containing the answer correctly with 79.3% accuracy; hence, the bulk of the difficulty lies in finding the exact span within the sentence.

|  | Exact Match | | F1 | |
| --- | --- | --- | --- | --- |
|  | Dev | Test | Dev | Test |
| Random Guess | 1.1% | 1.3% | 4.1% | 4.3% |
| Sliding Window | 13.2% | 12.5% | 20.2% | 19.7% |
| Sliding Win. + Dist. | 13.3% | 13.0% | 20.2% | 20.0% |
| Logistic Regression | 40.0% | 40.4% | 51.0% | 51.0% |
| Human | 80.3% | 77.0% | 90.5% | 86.8% |

Table 5: Performance of various methods and humans. Logistic regression outperforms the baselines, while there is still a significant gap between humans.

|  | $F_1$ | |
| --- | --- | --- |
|  | Train | Dev |
| Logistic Regression | 91.7% | 51.0% |
| – Lex., – Dep. Paths | 33.9% | 35.8% |
| – Lexicalized | 53.5% | 45.4% |
| – Dep. Paths | 91.4% | 46.4% |
| – Match. Word Freq. | 91.7% | 48.1% |
| – Span POS Tags | 91.7% | 49.7% |
| – Match. Bigram Freq. | 91.7% | 50.3% |
| – Constituent Label | 91.7% | 50.4% |
| – Lengths | 91.8% | 50.5% |
| – Span Word Freq. | 91.7% | 50.5% |
| – Root Match | 91.7% | 50.6% |

Table 6: Performance with feature ablations. We find that lexicalized and dependency tree path features are most important.

**Feature ablations.** In order to understand the features that are responsible for the performance of the logistic regression model, we perform a feature ablation where we remove one group of features from our model at a time. The results, shown in Table 6, indicate that lexicalized and dependency tree path features are most important. Comparing our analysis to the one in Chen et al. (2016), we note that the dependency tree path features play a much bigger role in our dataset. Additionally, we note that with lexicalized features, the model significantly overfits the training set; however, we found that increasing $L_2$ regularization hurts performance on the development set.

**Performance stratified by answer type.** To gain more insight into the performance of our logistic regression model, we report its performance across the answer types explored in Table 2. The results (shown in Table 7) show that the model performs best on dates and other numbers, categories for which there are usually only a few plausible candidates, and most answers are single tokens. The model is challenged more on other named entities (i.e., location, person and other entities) because there are many more plausible candidates. However, named entities are still relatively easy to identify by their POS tag features. The model performs worst on other answer types, which together form 47.6% of the dataset. Humans have exceptional performance on dates, numbers and all named entities. Their performance on other answer types degrades only slightly.

|  | Logistic Regression Dev F1 | Human Dev F1 |
| --- | --- | --- |
| Date | 72.1% | 93.9% |
| Other Numeric | 62.5% | 92.9% |
| Person | 56.2% | 95.4% |
| Location | 55.4% | 94.1% |
| Other Entity | 52.2% | 92.6% |
| Common Noun Phrase | 46.5% | 88.3% |
| Adjective Phrase | 37.9% | 86.8% |
| Verb Phrase | 31.2% | 82.4% |
| Clause | 34.3% | 84.5% |
| Other | 34.8% | 86.1% |

Table 7: Performance stratified by answer types. Logistic regression performs better on certain types of answers, namely numbers and entities. On the other hand, human performance is more uniform.

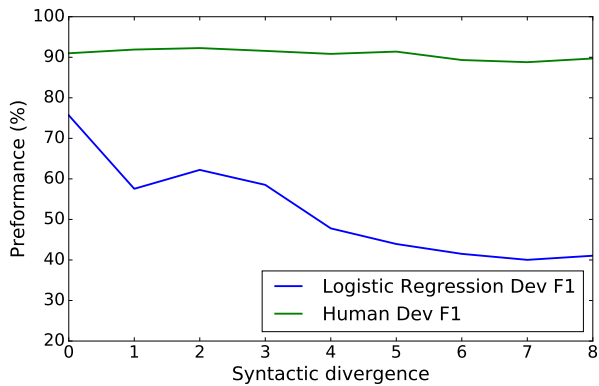

Figure 5: Performance stratified by syntactic divergence of questions and sentences. The performance of logistic regression degrades with increasing divergence. In contrast, human performance is stable across the full range of divergence.

**Performance stratified by syntactic divergence.**
As discussed in Section 4, another challenging aspect of the dataset is the syntactic divergence between the question and answer sentence. Figure 5 shows that the more divergence there is, the lower the performance of the logistic regression model. Interestingly, humans do not seem to be sensitive to syntactic divergence, suggesting that deep understanding is not distracted by superficial differences. Measuring the degree of degradation could therefore be useful in determining the extent to which a model is generalizing in the right way.

## 7 Conclusion

Towards the end goal of natural language understanding, we introduce the Stanford Question Answering Dataset, a large reading comprehension dataset on Wikipedia articles with crowdsourced question-answer pairs. SQuAD features a diverse range of question and answer types. The performance of our logistic regression model, with 51.0% F1, against the human F1 of 86.8% suggests ample opportunity for improvement. We have made our dataset freely available to encourage exploration of more expressive models. Since the release of our dataset, we have already seen considerable interest in building models on this dataset, and the gap between our logistic regression model and human performance has more than halved (Wang and Jiang, 2016). We expect that the remaining gap will be harder to close, but that such efforts will result in significant advances in reading comprehension.

## Reproducibility

All code, data, and experiments for this paper are available on the CodaLab platform:
https://worksheets.codalab.org/worksheets/0xd53d03a48ef64b329c16b9baf0f99b0c/ .

## Acknowledgments

We would like to thank Durim Morina and Professor Michael Bernstein for their help in crowdsourcing the collection of our dataset, both in terms of funding and technical support of the Daemo platform.